\documentclass[a4paper,11pt]{article}
\usepackage{fullpage}
\title{ {\bf Evaluating MT Systems: \\ A Theoretical Framework} }      
\author{
   {\bf Rajeev Sangal} \\
   IIIT Hyderabad\\
    {\tt sangal@iiit.ac.in}
  }

\date{}

\begin{document}
\maketitle

\begin{abstract}

This paper outlines a theoretical framework using which different automatic metrics 
can be designed for evaluation of Machine Translation systems. It introduces the concept
of {\em cognitive ease} which depends on {\em adequacy} and {\em lack of fluency}. 
Thus, cognitive ease becomes the main parameter to be measured rather 
than comprehensibility. The framework allows the components of cognitive ease to be
broken up and computed based on different linguistic levels etc. Independence of
dimensions and linearly combining them provides for a highly modular approach.

The paper places the existing automatic methods in an overall framework, to understand
them better and to improve upon them in future. It can also be used to evaluate the newer
types of MT systems, such as speech to speech translation and discourse translation..

\end{abstract}

{\small
\begin{center}
	\begin{minipage}{.8\textwidth}
\tableofcontents
		{\small
\input{mteval-framework-feb2022-sangal.toc}
                }
	\end{minipage}
\end{center}
}

\vspace{8mm}
Translation of a given ({\em source}) language text or spoken utterance into another language 
(called {\em target} language)
means preserving the meaning and generally creating the same effect on the reader of the
target language piece as the source language piece is intended to make on its reader.  
{\em Translation}, thus, is the
output in the target language, in the {\em spoken} or {\em written} mode,
and the unit might be a {\em sentence} or {\em text} or an {\em utterance}.

According to translation theorists, there is no such thing as a {\em perfect translation} which
is perfect for all. Suitability of a translation depends on the {\em purpose} and the {\em
audience}. For example, translation of an academic lecture should drop speech disfluencies,
whereas translation of a dialogue in a film expressing emotions should not. Similarly, while
translating technical terms from a source text, one should use {\em simplified terms} or {\em
exact terms} which might sound difficult, depending on whether the audience of the translation
consists of experts or common people.

\section{What to Evaluate in Translation}

The question is what to evaluate in a translation? The answer depends on
{\em for what} and {\em for whom} is the translation. Similarly, in the case of
machine translation, the answer depends on who is the consumer of the output translation
and for what the consumer is taking it? For example, is the translation for the end-user/reader 
or is it for the human translator/post-editor? Here are some possible purposes.

\subsection{Evaluating a Translation for the End User}

There is a need for evaluating a translation for the {\em end user} or {\em reader}, irrespective of
whether the translation is produced by human being or machine. The following are the parameters of
evaluation.
\begin{itemize}
	\item {\bf Comprehensibility} -- Is the reader able to comprehend the translation {\em correctly} and {\em easily}?
\end{itemize}
The above may be broken up into two parameters:
\begin{enumerate}
   \item {\bf Adequacy} -- Correctness or accuracy of translation
   \item {\bf Fluency} --  Ease of understanding or readability of translation by the reader
\end{enumerate}
The first one pertains to whether the information is given correctly in the translated sentence/text. 
The second one is about whether the reader is able to grasp the information, which is why 
fluency is important besides convenience to the reader. The former may be judged using
{\em information theoretic} aspects, whereas the latter is {\em cognitive}.

\subsection{Evaluating a Translation for the Human Translator}

Besides evaluation of a {\em translation}
for the end-user, there is also a need for
evaluating it for the {\em human translator}, who would be doing
post-editing etc., and producing the final translation for the end-user.
Following then are the issues:
\begin{itemize}
	\item {\bf Effort in comprehension}
	\item {\bf Effort in editing.}
\end{itemize}
The human translator has to first read the translated output (along with the source text);
and some effort is involved in {\em comprehending the meaning} of the text/sentence. Then he/she
has to {\em figure out the changes} that have to be made. All this is {\em cognitive effort}. Finally,
there is the {\em physical effort} in making changes to the given translation output using
a text editor or a translator's workbench or other tools. 

Physical effort of key clicks depends
on the kind of tool being used. So to make it neutral to any specific tool, it
would be estimated in terms of add-delete-replace operations and the time taken.
The major concern is 
the ease and speed of {\em producing the 
final output} by the human translator or post-editor (including the pre-editor, if any).

In the rest of the paper, when we say {\em cognition effort}, we would mean effort in {\em
comprehension} of the translation.\footnote{Effort in planning the edit changes, although cognitive, 
would not be estimated as it is too closely dependent on the kind of tools being used.}

Clearly, adequacy and fluency have a role in comprehension, but
the question is how do we estimate the cognition effort mentioned above using adequacy and fluency?

\subsection{Evaluating a Translation for System Development}

Here, the primary aim of the evaluation is to help the {\em developer} of the MT system. This would
require identifying the weakness in the MT system, and then suggesting ways to overcome it. It
normally requires either the analysis of the source or target sentences/text to identify {\em areas} where the
system is performing weakly, or the analysis of the weakness in the {\em approach} adopted to build
the MT system.

For example, in the case of rule-based systems, it usually requires linguistic analysis to say
on what sentence constructions or what vocabulary the system is performing poorly. With the
advent of stastical MT systems, choice of corpora for training became an important consideration,
and in the case of neural MT, attention and byte pair encoding are playing a major role.

\section{Cognitive Effort}

Estimation of the cognitive effort on part of the human translator can be in terms of
{\em comprehending} the translation, and in {\em planning the edits}.
As mentioned earlier, we will be ignoring the latter and focussing on only the comprehension in this paper.
It may be estimated based on adequacy and fluency at different {\em linguistic 
levels}, involving the parameters given below at the respective levels:
\begin{enumerate}
	\item {\bf Word Level} 
		\begin{itemize}
		   \item {\bf Content words:} Roots, prefixes, postfixes, terms, commonly used words/terms
		   \item {\bf Function words:} Post-positions, prepositions, inflections, derivations etc.
		\end{itemize}
	\item {\bf Chunk Level}\footnote{Chunks are simple or {\em non-recursive} phrases. They include 
		non-recursive noun phrases and verb groups. See also Local Word Groups in (Bharati et al., 1995).}
		\begin{itemize}
		   \item {\bf Intra-Phrase:} Correctness of words in a chunk,
			   number of chunks, chunk length
		   \item {\bf Inter-Phrases:} Number of chunks, named-entities in chunks,
			   commonly used named-entities, relations between chunks
		\end{itemize}
	\item {\bf Clause Level}
		\begin{itemize}
		   \item {\bf Intra-Clause:} Correctness of clause boundaries, 
				fragmentation\footnote{Fragmentation means dispersal of parts of a clause within a sentence, 
				                       which might be grammatically correct, but makes it difficult to read or comprehend.},
				vibhakti of chunks in a clause (ref Dependency Treebank for Hindi-Urdu)
			\item {\bf Inter-Clauses:} Number of clauses, long distance relationships among different clauses (Prasad, 2008), (Oza, 2009)
		\end{itemize}
	\item {\bf Discourse Level}
		\begin{itemize}
		   \item {\bf Cohesion:} Topic, focus, centering
		   \item {\bf Relations:} Implicit or explicit discourse relations
		\end{itemize}
		This level clearly works beyond the sentence level. In fact, how the clauses are configured
		is more important than the actual marking or delimitation of sentences.
\end{enumerate}
\label{sec:semantics_brief}
What about {\em semantics} and {\em pragmatics} of the translation? Semantics
relates directly with meaning and includes the following: entities, actions,   
relations between actions, flow of reasoning, etc.
The levels shown above deal with 
syntax (or sentence structure) and lexical items (or words), and indirectly capture 
the semantics.\footnote{For more on semantics, see Section~\ref{sec:semantics} 
later on in this paper.}
Pragmatics is not discussed in the paper (except for sentiment) as it is too difficult
to handle at the current stage of technology and understanding.
As mentioned earlier, {\em evaluation by automatic means} of the efforts in {\em comprehension} and {\em editing}
is the goal here. It is achieved by giving scores at the above levels and suitably combining them,
thus leading to automatic evaluation of the translated sentence.

\subsection{Score for Cognitive Ease}

We compute a score for {\em ease of cognition}\footnote{We use `ease' as opposed to `effort', so
that a higher score means better quality of translation.} of a given translated sentence/text. It is based
on the adequacy of information and the fluency of the sentence/text.

Let the score for {\em adequacy} be $A$, and the score for {\em lack of fluency} be $B$. Cognitive 
Ease $G$ is computed as,
    \[ G (A, B) = A * (1 - \gamma B^\delta) \]
simplified to,
    \[ G (A, B) = A * (1 - 0.5 * B) \]
The above indicates that there is information {\em contained} in the translation (scored by
$A$); however only a part of it might be grasped (or {\em cognized}) by the reader or the human
translator due to lack in fluency (scored by $B$).\footnote{There are issues regarding {\em
incorrect} information as opposed to correct but {\em incomplete} information in the translation,
discussed later in the paper (in Section~\ref{sec:perplexity}). Regarding information that is
present in the translation but missed in cognition, one has to recognize that the language has
redundancy. Therefore, part of the information that is missed from being cognized, hopefully
is obtained from other parts of the translation, or even surmised or inferred from other pieces
of information. This includes background information assumed to be with the reader/listener.}

The product of $A$ and $B$ shows that if adequacy or faithfulness to information content of
source is low (namely, score for $A$ is low), the overall cognition would naturally be low
(score for $G$ would be low). For example, if the information content is zero ($A=0$) then
no matter how fluent the sentence is, the net cognition (of what is said in the source sentence) 
would be $0$. On the contrary, if the
information content is faithful to the source ($A$ is $high$ or $1$), but it is presented with so
much disfluency ($B=1$) then the cognition is low (namely, $G$ is $low$).

The question is how does one compute the above, namely $G$, $A$, and $B$?  One method is to
compute the corresponding scores for each of the linguistic levels, and combine them to produce
the overall score. 

For the $i$th level, the cognitive score would be:
    \[ G_i (A_i, B_i)  = A_i * (1 - \gamma_i B_i^{\delta_i}) \]
Once the above are computed for each of the linguistic levels separately, how does one combine
them? Assuming {\em independence} of cognitive aspects at different levels, such scores can be combined
{\em linearly} to give the overall score $G$ for the translation:
    \[ G(A, B) = { \Sigma}_i  w_i * G_i (A_i, B_i) \]

\section{Cognition Scores at Different Structural Levels}

\subsection{Cognition Score at Word Level}

Major components at the word level which contribute to cognition are given below.
They are assumed to be independent, or in other words, they contribute to cognition
independent of each other. These contributions to cognition are added with suitable weights $\alpha$.

\subsubsection{Adequacy at Word Level ($A_1$)}

Score of adequacy at word level is given by $A_1$. This is how it can be broken into two parts, namely,
adequacy due to content words and due to function words; and their scores can be combined to yield the
score of $A_1$. For example,

\[ A_1 = \alpha_{11} P_{11} +  \alpha_{12} P_{12} \] 
where
      \begin{itemize}
	      \item $P_{11}$ is adequacy due to {\em lexical item} (root/stem) etc.
	      \item $P_{12}$ is adequacy due to {\em functional items} (function words,
		      case-endings or inflections, etc.).
         \item $\alpha$'s are the relative weights assigned to the above, therefore,
		 \[  \alpha_{11} +  \alpha_{12} = 1 \]
      \end{itemize}
Alternatively, $P_{11}$, $P_{12}$ may be written as $P_{1 lex}$,
               $P_{1 pos}$ respectively, for better readability. Thus, we would have,
	   \[ A_1 = \alpha_{1lex} P_{1lex} +  \alpha_{1pos} P_{1pos} \]
\label{sec:P_1lex}

If number of parameters at the word level is not just two, but $m_1$,  we would have:
	   \[ A_1 = { {\bf \Sigma}_{j=1}^{m_1} } ~\alpha_{1j} P_{1j} \]
where
	   \[ { {\bf \Sigma}_{j=1}^{m_1} } ~\alpha_{1j} = 1 \]

To compute the values of $P_{1j}$'s (namely, the adequacy parameters), one would have to align
the candidate translated sentence with the reference translated sentence, and compute the match 
or similarities between the words so aligned, much like Meteor (Banerjee et al., 2005).\footnote{Matching may 
be done using exact match between words, or matching the stems/roots, or using synonyms.}

\subsubsection{Lack of Fluency at Word Level ($B_1$)}

The words in the sentence may pose difficulties in reading, such as:

\[ B_1 = \beta_{11} Q_{11} +  \beta_{12} Q_{12} +  \beta_{13} Q_{13} \] where
      \begin{itemize}
	      \item $Q_{11}$ is lack of fluency due to (large) {\em number of words},
	 \item $Q_{12}$ is lack of fluency due to {\em uncommon} words\\
		 (approximated by the number of less frequent words in the sentence).\footnote{New 
		      entities (nouns) introduced in a sentence, which have not occurred in the previous sentences,
		      shows difficulty of fluency. However, these are taken into account at the discourse
		      level and not at the word level.}
	      \item $Q_{13}$ is lack of fluency due to {\em terms},

         \item $\beta$'s are the relative weights assigned to the above, therefore,
		 \[  \beta_{11} +  \beta_{12} +  \beta_{13} = 1 \]
      \end{itemize}
Alternatively, $Q_{11}$, $Q_{12}$, $Q_{13}$ may be written as $Q_{1 \#word}$,
	       $Q_{1uncom}$, $Q_{1term}$ respectively, for better readability.
Thus, we would have,
	   \[ B_1 = \beta_{1\#word} Q_{1\#word}  + \beta_{1uncom} Q_{1uncom}   + \beta_{1term} Q_{1term} \]

Values of $\beta$'s can be determined empirically by conducting experiments on
cognitive ease. These may even be tailor made to a given situation or type of audience.  
For example, if the MT output in a technical domain is being read by people
unfamiliar with the domain, a higher value $\beta_{1term}$ would be in order.

If number of parameters at the word level is not just three, but $n_1$,  we would have:
	   \[ B_1 = { {\bf \Sigma}_{j=1}^{n_1} } \beta_{1j} Q_{1j} \]
where
	   \[ { {\bf \Sigma}_{j=1}^{n_1} } \beta_{1j} = 1 \]

\subsubsection{Score at Word Level ($G_1$)}

Formula for ease of cognition then, at the word level, is,
    \[ G_1 = A_1 * (1 - \gamma_1 * B_1^{{\delta}_1} ) \]
approximated by its simplified form:
    \[ G_1 = A_1 * (1 - 0.5 * B_1) \]
The above captures the intuition that although adequacy score ($A_1$) might be
what it is, if the fluency is lacking or bad, it reduces cognitive ease in grasping the meaning.
Thus, a higher value of $B_1$ leads to a lower value of $G_1$.
Alternatively, if the information content in the translation ($A_1$) is low, then no matter how
fluent the translation is, the overall score is low.

\subsection{Cognition Score at Chunk Level}

\subsubsection{Adequacy at Chunk Level ($A_2$)}
    \[ A_2 = \alpha_{21} P_{21} +  \alpha_{22} P_{22} \] 
where
      \begin{itemize}
	      \item $P_{21}$ is adequacy due to content words in {\em chunks}, which 
		      covers {\em correctness} of content words particularly {\em heads} of chunk,
	      \item $P_{22}$ is adequacy due to {\em vibhakti} or function markers with the
		      chunk.\footnote{The term vibhakti includes prepositions in English or post-positions in
		      Hindi, or case-endings in Dravidian languages, and so on.
		      {\em Generalized vibhakti} (Bharati et al., 1997)
		      also includes order (or subject and object in English), but that is kept
		      out of bounds because at the chunk level such information is not available.}
              \item $\alpha$'s are the relative weights assigned to the above totaling to $1$.
      \end{itemize}

\subsubsection{Lack of Fluency at Chunk Level ($B_2$)}
    \[ B_2 = \beta_{21} Q_{21} +  \beta_{22} Q_{22} +  \beta_{23} Q_{23} \]
where
      \begin{itemize}
	      \item $Q_{21}$ is lack of fluency due to {\em large number} of words in chunks (intra-chunk property)
	      \item $Q_{22}$ is lack of fluency due to {\em large number} of chunks
	      \item $Q_{23}$ is lack of fluency due to {\em uncommon named-entities}
              \item $\beta$'s are the relative weights assigned to the above totaling to $1$.
      \end{itemize}

\subsubsection{Score at Chunk Level ($G_2$)}

Formula for the ease of cognition at the chunk level is similar to that at word level,
    \[ G_2 = A_2 * (1 - \gamma_2 * B_2^{{\delta}_2} ) \]
and is approximated by its simplified form:
    \[ G_2 = A_2 * (1 - 0.5 * B_2) \]

\subsection{Cognition Score at Clause Level}

\subsubsection{Adequacy at Clause Level ($A_3$)}
    \[ A_3 = \alpha_{31} P_{31} +  \alpha_{32} P_{32} \] 
where
      \begin{itemize}
	      \item $P_{31}$ is adequacy due to chunks in {\em clauses}, which covers
		      correctness of chunks relationships among the chunks within each
		      clause (intra-clause property),
	      \item $P_{32}$ is adequacy due to correctness of {\em relationship} among {\em clauses},
              \item $\alpha$'s are the relative weights assigned to the above totaling to $1$.
      \end{itemize}

\subsubsection{Lack of Fluency at Clause Level ($B_3$)}

    \[ B_3 = \beta_{31} Q_{31} +  \beta_{32} Q_{32} +  \beta_{33} Q_{33} \]
where
      \begin{itemize}
	      \item $Q_{31}$ is lack of fluency due to {\em large number} of chunks within each 
		      clause (intra-clausal property)
	      \item $Q_{32}$ is lack of fluency due to {\em fragmentation} of each clause
	      \item $Q_{33}$ is lack of fluency due to {\em long distance relationships} among clauses
              \item $\beta$'s are the relative weights assigned to the above totaling to $1$.
      \end{itemize}

\subsubsection{Score at Clause Level ($G_3$)}

Formula for the ease of cognition at the clause level is,
    \[ G_3 = A_3 * (1 - \gamma_3 * B_3^{{\delta}_3} ) \]
approximated by its simplified form:
    \[ G_3 = A_3 * (1 - 0.5 * B_3) \]

\subsection{Cognition Score at Discourse Level}

\subsubsection{Adequacy at Discourse Level ($A_4$)}

    \[ A_4 = \alpha_{41} P_{41} +  \alpha_{42} P_{42} \] 
where
      \begin{itemize}
	      \item $P_{41}$ is adequacy due to {\em topic} and {\em focus}\footnote{Alternatively, the
		      notion of {\em centering} may be used from the centering theory (Grosz et. al, 1995).}
	      \item $P_{42}$ is adequacy due to correct {\em discourse relations} across clauses/sentences
              \item $\alpha$'s are the relative weights assigned to the above totaling to $1$.
      \end{itemize}

\subsubsection{Lack of Fluency at Discourse Level ($B_4$)}

    \[ B_4 = \beta_{41} Q_{41} \]
where
      \begin{itemize}
	      \item $Q_{41}$ long distance relations between linked clauses,
	      \item $\beta_{41}$ is equal to 1.   
      \end{itemize}

\subsubsection{Score at Discourse Level ($G_4$)}

Formula for the ease of cognition at the discourse level,
    \[ G_4 = A_4 * (1 - \gamma_4 * B_4^{{\delta}_4} ) \]
approximated by its simplified form:
    \[ G_4 = A_4 * (1 - 0.5 * B_4) \]

\subsection{Summary of Scoring Formulas}
\label{sec:summ_scoring}

The scoring formulas for {\em adequacy} $A$ and {\em lack of fluency} $B$ respectively, at level $i$ can be written compactly as:
\begin{center}
	          \( A_i = { {\bf \Sigma}_{j=1}^{m_i} } \alpha_{ij} P_{ij} \)  ~~and~~  \( { {\bf \Sigma}_{j=1}^{m_i} } \alpha_{ij} = 1 \)\\
\end{center}
		  and
\begin{center}
	          \( B_i = { {\bf \Sigma}_{j=1}^{n_i} } \beta_{ij} Q_{ij} \)  ~~and~~  \( { {\bf \Sigma}_{j=1}^{n_i} } \beta_{ij} = 1 \)
\end{center}
Thus, there would be one equation for $A_i$ and one for $B_i$ for every value of $i$ from $1$ to $L$, resulting in $2*L$ equations.
$L$ is the number of levels (or dimensions).

{\em Cognitive Ease} $G_i$ at level $i$ is given by:
    \[ G_i (A_i, B_i)  = A_i * (1 - \gamma_i B_i^{\delta_i}) \]
and the overall {\em cognitive ease} aggregated over all values of $i$ is given by:
    \[ G(A, B) = {\Sigma}_{i=1}^L  w_i * G_i (A_i, B_i) \]
The values of the weights $\alpha_{ij}$, $\beta_{ij}$ and $w_i$ for all values of $i,j$ are arrived at using experiments.

\section{Cognition Score Based on Semantics}
\label{sec:semantics} 

As mentioned in Section~\ref{sec:semantics_brief} (on page~\pageref{sec:semantics_brief}), semantics pertains to
entities, actions, flow of reasoning, etc. These are detailed out below.

\begin{enumerate}
	\item {\bf Meaning of words}
		\begin{itemize}
		   \item {\bf Words and terms:} Words or terms occurring in the translation.
		   \item {\bf Semantic distance:} Distance between meaning of a word in 
			   the source text and the meaning of the target word in the  translation. (In case,
				a reference translation is available, then it could be computed using the {\em semantic
				distance} between the word in the candidate translation and the corresponding word in 
				the reference. A synonym dictionary or a thesaurus or Wordnet can be of help here.)
		\end{itemize}
	\item {\bf Entities}
		\begin{itemize}
		   \item {\bf Entities introduced:} Correctness of entities expressed in the target text,
                         and ease of cognizing them
		   \item {\bf Relations between entities:} Relations marked between entities, continuity of 
			   referred entities in the discourse (spoken or written)
		\end{itemize}
		These would clearly be closest to the chunk level (level 2) described earlier.
	\item {\bf Actions:} Relations between entities and action (or verb), and relations among
                 the actions.
		\begin{itemize}
		   \item {\bf Actions introduced:} Correctness of each action structure, that is, action and the
			   entities participating in the action (as expressed by dependency or similar
				language structure) and ease of cognizing it
		   \item {\bf Relations among actions:} Relations between actions
		\end{itemize}
		These would be closest to the clause level (level 3) described earlier,
		where typically the noun chunks are related to the verb chunks.
	\item {\bf Flow of entities:} The sequence in which the entities are introduced, known
		entities to unknown or newly introduced entities, etc. (This relates with the {\em topic}
		and {\em focus} at the discourse 
		level (level 4), which hopefully captures the introduction of certain saliant
		entities.)\footnote{The sentence boundaries may be taken liberty with, in translation, as long 
		as the flow is fine. Of course, the evaluation would have to be at the paragraph
		level and not at individual sentence level.}
	\item {\bf Flow of reasoning:} The flow of reasoning or buildup of the argument (say, described using
		Rhetorical Structure Theory (ref RST)). This would bring up relations within a discourse 
		segment. (Similarly, inter-segment relations could also be considered at another level.)
	\item {\bf Sentiment:} Positive or negative sentiments expressed for the entities or 
		actions. This expression is done by choice of words as well as particles etc.
		This clearly needs to be worked out at the word level and also at chunk level.
\end{enumerate}
Note that {\em continuity of information} as obtained from discourse has been included. However,
{\em background information} has been ignored in the above. This is because of the difficulty
of modeling the latter. Different listeners would have different background
knowledge, and modeling them for each one would be extremely difficult. Limited experiments in
background knowledge could be conducted for people having subject domain knowledge and those not
having it.

The question is how to take the scores due to the above semantic parameters and combine them with 
scores from different structural levels. Those semantic parameters which are closely associated with
or have a close correspondence with the structural parameters are better left out because
of two reasons: 
\begin{enumerate}
	\item Tools for computing the semantic parameters are generally less
             developed than those for structural parameters, and as a result are more error prone.
        \item Those semantic parameters which are related with the structural parameters would
		no longer be {\em independent} if both are included, as they would violate the
		independence assumption.

              Note that \label{sec:score_linear} the independence of different levels in their
	      contribution to cognition and therefore combining their score by a {\em linear combination} over $i$ levels:
		 \[ G(A, B) = {\Sigma}_i w_i * G_i (A_i, B_i) \]
              is a simplifying assumption which may not be quite true.
	      It would need to be refined in due course by identifying
              those elements which have an inter-dependence on each other. Such parts can be
              brought together into {\em orthogonal dimensions} whose scores can then
	      be linearly combined.\footnote{What is said regarding independence of $G_i$'s 
		in their contribution to $G$ also applies to independence 
		of $P_{ij}$'s and $Q_{ij}$'s for all values of $j$ in their contribution to $A_i$ and $B_i$
		respectively at each level $i$.  For example, if
		$P_{11}$, $P_{12}$, etc. are independent of each other in their contribution to $A_1$, 
		they can be combined linearly, and similarly if
		$Q_{11}$, $Q_{12}$, etc. are independent of each other in their contribution to $B_1$, 
		they can be combined linearly.}
\end{enumerate}

Awaiting experiments regarding what parameters from semantics should be chosen and how they
should be combined with which structural parameters, our guess is that in case of speech to speech
translation, the {\em flow of entities} is extremely important. Topic and focus capture the
important aspects of the flow, but in the case of oral communication the flow in itself is also very
important. It might also relate with the speech {\em utterance unit} which is the counterpart
of sentence in written text. Entities and prosody features might help in the identification of
the utterance unit.

\section{Adding New Dimensions/Levels}

The formulas described above, make it easy to add new independent dimension/level or modify the
formulas for the existing dimensions/levels. These are illustrated by introducing elements of semantics.

\subsection{Incorporating Score for a New Dimension}

\label{sec:flow_entities}

Suppose we wish to add the dimension of {\em flow of entities} from semantics.
To incorporate the flow of entities (which is a part of semantic dimension) along with the {\em linguistic
structural levels} in the evaluation, the following can be done: Introduce another {\em orthogonal
dimension}, say level 5, for the flow of entities. Since it is orthogonal to the other four dimensions, the change is easy. Simply
call it $G_5$, and introduce its adequacy ($A_5$) and lack of fluency ($B_5$) :\\
    \( ~~~~~~~~~~~~A_5 = \alpha_{51} P_{51} \)\\
where
      \begin{itemize}
	      \item $P_{51}$ is adequacy in {\em sequence of entities},
	      \item $\alpha_{51}$ is the weight equal to $1$, because there is only one term in $A_5$.
      \end{itemize}
Lack of fluency in the flow of entities is given by:\\
    \( ~~~~~~~~~~~~B_5 = \beta_{51} Q_{51} \)\\
where
      \begin{itemize}
	      \item $Q_{51}$ is {\em number of entities} and their {\em sequence},\footnote{When there is multi-modal
		     input, say, there is video with the audio, the lack of correspondence between 
		     the translated audio and original video might lead to a further lack of fluency.
		      For example, the gesture in the video might point to an entity, but the 
		      translated audio output might have a phrase referring to another entity. Lip
		      synchronization is another common aspect while putting the translated audio in video.}
	      \item $\beta_{51}$ is the weight equal to $1$, because there is only one term in $B_5$.
      \end{itemize}
(In case there were more parameters 
besides $P_{51}$, they could be introduced here as well, say, as $P_{52}$ and $\alpha_{52}$, etc.)

Now, {\em Cognitive Ease} $G_5$ at level $5$ is given by:
    \[ G_5 (A_5, B_5)  = A_5 * (1 - \gamma_5 B_5^{\delta_5}) \]
and the overall {\em Cognitive Ease} aggregated over all values of $i$ is given by:
    \[ G(A, B) = {\Sigma}_{i=1}^{5}  w_i * G_i (A_i, B_i) \]
Thus, introducing a new orthogonal dimension is a modular process which does not affect
other levels. The only thing that needs to be done is to change the relative weights $w_i$'s suitably in
the final equation above.

\subsection{Making Changes in an Existing Dimension}

Making changes in an existing level can also be done. If 
another parameter is to be introduced, it can be done easily provided the parameter
is orthogonal to other parameters at the same level. For example, if the {\em sentiment} has to 
be introduced at the word level (level 1), it can again be done modularly by adding additional terms
of $P_{13}$ and $Q_{14}$, shown in bold below:\\ \\
\( ~~~~~~~~~~~A_1 = \alpha_{11} P_{11} +  \alpha_{12} P_{12} \)  + \boldmath  \( \alpha_{13}  P_{13}  \) \\ \unboldmath \\
\( ~~~~~~~~~~~B_1 = \beta_{11} Q_{11} +  \beta_{12} Q_{12} + \beta_{13} Q_{13} \) + \boldmath \( \beta_{14} Q_{14}  \) \\ \unboldmath
where
      \begin{itemize}
	      \item {\bf $P_{13}$ } is adequacy regarding {\em sentiment}
	      \item {\bf $Q_{14}$ } is lack of fluency due to ambiguity, if any, in sentiments
      \end{itemize}
Nothing else changes in the formula of $G_1$. Similarly, there is no change in the formulas 
of the other $G_i$'s, and in the overall formula of $G$.

\subsection{Incorrect Information in Translation}
\label{sec:perplexity}

There are two types of errors that might be there in the translation: (1) full information in the
source is not provided in the translation, although no wrong information is given,
and (2) wrong information is provided in transalation. We know that there is redundancy in language.
So even if information at one point is missed, hopefully it is conveyed at another point. This
will be particularly true as we move from evaluation of translation of isolated sentences to
paragraphs and whole texts.

Question is how do we deal with wrong information in translation. This might affect the
comprehensibility scores much more. It is proposed that in future {\em perplexity} scores may be 
computed and made a part of lack of fluency.

\section{Computing Adequacy and Lack of Fluency Parameters}

The next question is how to compute the parameters $P_{ij}$'s and $Q_{ij}$'s? Their
computation requires linguistic features, which could be obtained using linguistic tools, 
statistical data, and even neural models. We will discuss these next.

\subsection{Computing Parameters at Word Level}

\subsubsection{Computing Adequacy Parameters at Word Level ($P_{1j}$)}

Adequacy parameters at the word level are $P_{1j}$'s. As mentioned earlier, one would have
to first align the candidate translation (output from the MT system being evaluated) with the
reference translation. 
It would require making either exact word match, or root/stem match or synonym match.

Once the matches are obtained, value of the parameter $P_{11}$ can be
computed using precision and recall of words in the candidate translation
as compared with the reference translation, like in Meteor.
It would mean finding the distance between the two translations to
determine adequacy. This could be done by counting the number of matches
out of the total number of words to determine precision and recall and then
F-mean or $P_{11}$. 
   \[ Prec = \#words\_matched\_candidate / \#words\_candidate   \]
   \[ Recall = \#words\_matched\_candidate / \#words\_ref   \]
and,
   \[ P_{11} = 10 * Prec * Recall / (Recall + 9 * Prec) \]
where,
\begin{itemize}
	\item $\#words\_matched\_candidate$  =\\ Number of words that matched between candidate
		sentence and reference sentence 
	\item $\#words\_candidate$  = Total number of words in the candidate sentence
	\item $\#words\_ref$  = Total number of words in the reference sentence
\end{itemize}
Finally,
	   \[ A_1 = \alpha_{11} P_{11} \]
As an example here, $P_{11}$ based on words is shown. However if necessary, 
parameters different from the above, such as $P_{1lex}$ and  $P_{1pos}$ 
could be used (see Section~\ref{sec:P_1lex}) instead of words. 
$\alpha_{1lex}$ and $\alpha_{1pos}$ would have to be determined empirically.

Meteor gives a good method for the computation of the score at the word level.  However, with the
word level score, it introduces a penalty based on lesser number of chunks, thus mixing the two levels. Even though it
gives good results, it comes in the way of building a more modular and ultimately a more accurate 
metric. Therefore, we recommend that the two levels be kept separate.

\subsubsection{Computing Fluency Parameters at Word Level ($Q_{1j}$)}

Similarly, as the number of {\em words} and number of {\em uncommon} words in the
candidate sentence affect fluency, these could be found, and based on these the 
sentence may be given a lack of fluency score. 
This could be compared with the average length of the sentence
in the language. (It might even be independent of the reference sentence.) Thus,
the computation might look like the following:
\begin{itemize}
	\item[] $Q_{11}$ = \#words / \#ave\_sentence\_len
	\item[] $Q_{12}$ = \#uncommon\_words / \#ave\_sentence\_len
\end{itemize}
and,
   \[ B_1 = \beta_{11} Q_{11}  + \beta_{12} Q_{12} \]
$\beta_{11}$ and $\beta_{12}$ would have to be determined empirically.

\subsection{Computing Parameters at Chunk Level}

The chunks in the candidate sentence and the reference sentence would have to be aligned before
computing the parameters. Once aligned, one can compare the respective aligned chunks.

\subsubsection{Computing Adequacy Parameters at Chunk Level ($P_{2j}$)}

At the chunk level, we can compute the correctness of head words and function words.
      \begin{itemize}
	      \item $P_{21}$ is adequacy 
		      related to the correctness of the {\em head word} in the chunks in the candidate sentence
		      as compared with the aligned chunks in the reference sentence,
	      \item $P_{22}$ is adequacy due to {\em vibhakti} or function markers with the
		      two aligned chunks.
      \end{itemize}
Computation of $P_{21}$ and $P_{22}$ based on precision and recall is as follows:
\begin{itemize}
   \item Prec\_content =  \#chunks with correct heads / \#chunks in candidate sentence
   \item Recall\_content =  \#chunks with correct heads / \#chunks in reference sentence
\end{itemize}
and,
\[ P_{21} = 10 * Prec\_content * Recall\_content / (Recall\_content + 9 * Prec\_content) \]
Similarly, one can compute based on the correct number of function words,
\begin{itemize}
    \item $Prec\_func\_word$ =  \#chunks with correct function words / \#chunks in candidate sentence
    \item $Recall\_func\_word$ =  \#chunks with correct function words / \#chunks in reference sentence
\end{itemize}
and,
\[ P_{22} = 10 * Prec\_func\_word * Recall\_func\_word / (Recall\_func\_word + 9 * Prec\_func\_word) \]
Finally,
    \[ A_2 = \alpha_{21} P_{21} +  \alpha_{22} P_{22} \] 
where, $\alpha$'s are determined empirically.

\subsubsection{Computing Fluency Parameters at Chunk Level ($Q_{2j}$)}

If the number of chunks in the candidate sentence is the sole criterion for lack of fluency, it can be computed
as follows:
\begin{itemize}
	\item  $Q_{21}$ = \#chunks in candidate sentence / \#ave num of chunks in a sentence in the language
\end{itemize}
and,
   \[ B_2 = \beta_{21} Q_{21}  \]
where, $\beta_{21}$ is determined empirically.

\vspace{2mm}
Scores at clause and discourse level can be computed similarly, and are not given here.

\subsection{Computing Parameters at Semantic Level}

Parameters may be similarly computed at clause/sentence level, and discourse level. Here we look at the semantic level.

If we include the sequence of entities (see Section~\ref{sec:flow_entities}) as they occur in the source sentence, comparision can be done
with the sequence in the target candidate sentence. Scoring can be done as follows, for example:
   \[ P_{51} = Compare\-Length (seq\_entities\_source, seq\_entities\_candidate)  \]
   \[ P_{52} = Edit\-Dist (seq\_entities\_source, seq\_entities\_candidate) \]
where,
\begin{itemize}
	\item[] $Compare\-Length$ = Function that compares lengths but gives a step-wise score\\
		(For example, equal lengths get a score of 1, if the difference in the lengths is within 20\%, it gets a score 
		of 0.9; with difference of 30\%, it gets the score of 0.75, and with less than 50\%
		it gets a score of 0.)
	\item[] $Edit\-Dist$ = Edit distance based on number of operations of add-delete-modify
\end{itemize}
and,
    \[ A_5 = \alpha_{51} P_{51} +  \alpha_{52} P_{52} \]

Computation of lack of fluency parameter $B_{5j}$ is ignored because it is already taken care of at the
chunk level. Thus, $B_{5} = 0$.

\section{Conclusions}

We have tried to look at translations from the perspective of linguistics and cognition.
It would help us gain clarity about the human judgements regarding translated sentence, such as
what makes adequacy and fluency. Four linguistic levels, from words to discourse (paragraph),
are independent to an extent, and therefore they can be computed separately and combined linearly.

The existing automatic methods of evaluation such as Bleu (Papineni et al., 2002), 
Meteor (Banerjee et al., 2005) (Gupta et al., 2009), Comet (Rei et al., 2020) can be placed within the
framework, to understand them better. In this sense, it also helps us in designing more accurate
automatic metrics which are closer to human judgements.  It is also hoped that the metrics can
change and improve continuously because we understand the parameters affecting cognitive ease.
The metrics can of course be tuned to the availability of language analysis tools, given a 
target language.

In future, editing ease can also be incorporated besides the comprehension ease covered in this
paper. This would include the planning of editing which is a cognitive effort, and then actually
making changes by performing the keystrokes etc.

Stability of the MT system is another concern. A seemingly well performing system under one kind
of evaluation, suddenly starts performing badly with a change in subject domain or genre.
What this means is that in real-life, users are thrown off by the fluctuation in system 
performance whenever there is a change in input texts etc.
Stability needs to be studied, and requires a theoretical framework for the study.

The evaluation framework we hope will inspire researchers to design specific metrics for their
tasks and concerns. By properly
designed metrics in the framework, it might also help the developers to focus on those
aspects of their MT system, where the score is lower. Thus, it might directly help in building
better MT systems. It can also open up the evaluation of new types of
MT systems; for example, if discourse translation as well as multi-modality are to be taken
into consideration, newer metric may be defined guided by the framework.

\section*{References}

\begin{enumerate}
	\item Banerjee, Satanjeev and Alon Lavie, ``METEOR: An Automatic Metric for MT Evaluation with Improved Correlation
		with Human Judgments", In {\em Proc. of the ACL 2005, Workshop on Intrinsic and Extrinsic Evaluation Measures
		for MT and/or Summarization}, ACL, 2005.
	\item Begum, Rafiya, Samar Husain, Arun Dhwaj, Dipti Misra Sharma, Lakshmi Bai and Rajeev Sangal,
		``Dependency Annotation Scheme for Indian Languages",
		In {\em Proc. of The Third International Joint Conference on Natural Language
		Processing (IJCNLP)}, ACL anthology, 2008.
	\item Bharati, Akshar, Rajni Moona, Smriti Singh, Rajeev Sangal, and Dipti
		Misra Sharma, ``MTeval: An Evaluation Methodology for Machine
		Translation Systems", {\em Proc. of SIMPLE-04: Symposium on Indian
		Morphology, Phonology \& Language Engineering}, IIT Kharagpur, 2004
	\item Bharati, Akshar, Vineet Chaitanya and Rajeev Sangal, {\em Natural
		Language Processing: A Paninian Perspective}, Prentice-Hall of India,
		New Delhi, 1995.
	\item Bharati, Akshar, Medhavi Bhatia, Vineet Chaitanya and Rajeev Sangal,
		``Paninian Grammar Framework Applied to English", {\em South Asian Langauge
		Review} journal, Creative Books, New Delhi, 1997.
	\item Doddington, George, ``Automatic Evaluation of Machine
		Translation Quality Using N-gram Co-occurence Statistics",
		In {\em Human Language Technology: Notebook Proc.}, ACL,
		pages 128-132
	\item Grosz, Barbara J., Aravind K. Joshi, and Scott Weinstein,
		``Centering: A Framework for Modeling the Local Coherence of Discourse",
		{\em Computational Linguistics}, Volume 21, Number 2, ACL, June 1995.
	\item Gupta, Ankush, Sriram Venkatapathy, and Rajeev Sangal
		``METEOR-Hindi : Automatic MT Evaluation Metric for Hindi as a Target
		Language", {\em International Conference on Natural Language Processing (ICON)}, 
		Hyderabad, Dec 2009.
	\item Mann, William C., Sandra A. Thompson, ``Rhetorical Structure Theory: Toward a Functional Theory of Text Organization",
		{\em Text: Interdisciplinary Journal for the Study of Discourse}, 8 (3), 1988, pp243-281.
	\item Oza, Umangi, Rashmi Prasad, Sudheer Kolachina, Dipti Misra Sharma, and Aravind Joshi,
		``The Hindi Discourse Relation Bank", ACL Anthology, 2009.
	\item Papineni, Kishore, Salim Roukos, Todd Ward, and Wei-Jing
		Zhu, ``BLEU: A Method for Automatic Evaluation of
		Machine Translation", In {\em Proc. of the 40th Annual
		Meeting of the Association for Computational Linguistics}, ACL, 2002.
	\item Prasad, Rashmi, Nikhil Dinesh, Alan Lee, Eleni Miltsakaki, Livio Robaldo, Aravind Joshi, and Bonnie Webber,
		``The Penn Discourse TreeBank 2.0", {\em Proc. of the Sixth International Conference on Language 
		Resources and Evaluation (LREC'08)}, ELRA, ACL, 2008.
	\item Rei, Ricardo, Craig Stewart, Ana C Farinha, and Alon Lavie,
		``COMET: A Neural Framework for MT Evaluation",
		{\em Proc. of the 2020 Conference on Empirical Methods in Natural Language Processing (EMNLP)}, ACL, 2020.
\end{enumerate}

\vspace{4mm}
\noindent
{\em 7 February 2022}

\end{document}